\documentclass{article}

\usepackage{algorithm2e}
\SetKwRepeat{Do}{do}{while}
\usepackage{amsfonts}
\usepackage{amsmath}
\usepackage{bm}
\usepackage{booktabs}
\usepackage{mathtools}
\usepackage{pgfplots}
\usepackage{tikz}
\usepackage{url}
\usetikzlibrary{arrows, backgrounds}
\usetikzlibrary{external}
\tikzstyle{arrow} = [thick,->,>=stealth, line width=0.8pt]
\DeclareMathOperator*{\argmax}{arg\,max}
\DeclareMathOperator*{\argmin}{arg\,min}
\DeclareMathOperator{\E}{\mathbb{E}}
\DeclareMathOperator{\R}{\mathbb{R}}
\pgfplotsset{
    jittery/.style={
        y filter/.code={\pgfmathparse{\pgfmathresult+rnd*#1}}
    },
    jittery/.default=0.1
}
\pgfplotsset{
    jitter/.style={
        x filter/.code={\pgfmathparse{\pgfmathresult+rnd*#1}}
    },
    jitter/.default=0.1
}
\pgfplotsset{every axis/.append style={
  label style={font=\small},
  tick label style={font=\small},
  xlabel near ticks,
  ylabel near ticks,
}}
\pgfplotsset{every axis plot post/.append style={
  only marks,
  jitter=0.02,
}}

\title{\Large \bf Hidden Markov Random Field Iterative Closest Point}
\author{John Stechschulte and Christoffer Heckman$^\ast$%
\thanks{All authors are with the Autonomous Robotics and
Perception Group at the University of Colorado, Boulder.}%
\thanks{$^\ast$ Corresponding author. E-mail: {\tt\small
christoffer . heckman at colorado.edu}}} %chktex 12 chktex 26
\date{5 June 2017}

\begin{document}

\newlength\figureheight%
\newlength\figurewidth%

\maketitle
\thispagestyle{empty}
\pagestyle{empty}

\begin{abstract}
  When registering point clouds resolved from an underlying \mbox{2-D} pixel
  structure, such as those resulting from structured light and flash LiDAR
  sensors, or stereo reconstruction, it is expected that some points in one
  cloud do not have corresponding points in the other cloud, and that these
  would occur together, such as along an edge of the depth map. In this work, a
  hidden Markov random field model is used to capture this prior within the
  framework of the iterative closest point algorithm. The EM algorithm is used
  to estimate the distribution parameters and the hidden component memberships.
  Experiments are presented demonstrating that this method outperforms several
  other outlier rejection methods when the point clouds have low or moderate
  overlap.
\end{abstract}

\section{Introduction}
Depth sensing is an increasingly ubiquitous technique for robotics, \mbox{3-D}
modeling and mapping applications. For indoor scenes, structured light depth
sensors are very popular and LiDAR is frequently used for robotics applications
outdoors, including for autonomous vehicles. Both of these sensors output point
clouds, or points on surfaces in \mbox{3-D} space. An algorithm to register
point clouds is essential to making use of the resulting data from these popular
mobile sensors. Iterative closest point (ICP)~\cite{icp_besl} is commonly used
for this purpose, although this time-tested approach has its drawbacks,
including failure when joining and transforming non-overlapping
clouds~\cite{makadia2006fully}~\cite{trimmedicp}. In fact, many applications of
point cloud registration are specifically intended to have non-overlapping point
clouds, such as combining several partial scans into a complete \mbox{3-D}
model. To address this gap, we present a probabilistic model using a hidden
Markov random field (HMRF) for inferring which points in a cloud lie in the
overlap, and use the EM algorithm to carry out this inference.

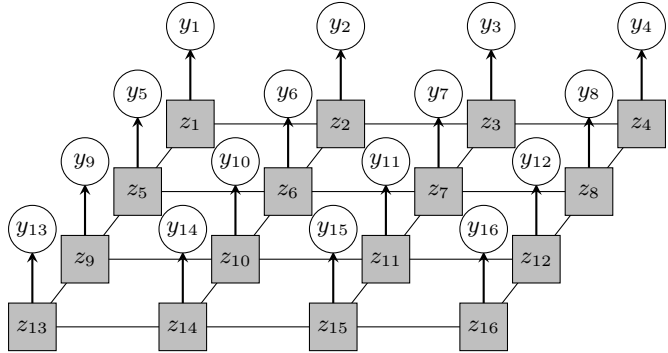
\begin{figure}[t]\begin{center}
  \begin{tikzpicture}[block/.style={
        draw,
        fill=white,
        minimum width=18pt,
        minimum height=18pt,
        font=\small,
        inner sep=0
      }]
      \foreach \i in {0,...,3} {
        \begin{scope}[on background layer]
          \draw (0.35*2*\i, 0.45*2*\i) -- (6+0.35*2*\i, 0.45*2*\i);
          \draw (2*\i, 0) -- (2*\i+6*0.35, 6*0.45);
        \end{scope}
        \foreach \j in {0,...,3} {
          \pgfmathtruncatemacro{\label}{4*(3-\j)+\i+1}
          \draw [arrow] (2*\i+2*\j*0.35, 2*\j*0.45) --
              (2*\i+2*\j*0.35, 1+2*\j*0.45);
          \node [block, fill=lightgray] at (2*\i+2*\j*0.35, 2*\j*0.45)
              {$z_{\label}$};
          \node [block, circle] at (2*\i+2*\j*0.35, 1.3+2*\j*0.45)
              {$y_{\label}$};
        }
      }
  \end{tikzpicture}
  \caption{Graphical model for nearest fixed point distance, shown for a
  $4\times4$ grid of pixels in the free depth map. At pixel $i$, the $Z_i \in
  \{\pm 1\}$ value is the unobserved inlier/outlier state, and $Y_i \in
  \R_{\geq 0}$ is the observed distance to the closest fixed point.}\label{pgm}
\end{center}\end{figure}

\begin{figure}
  \begin{center}
    \includegraphics[width=0.9\textwidth]{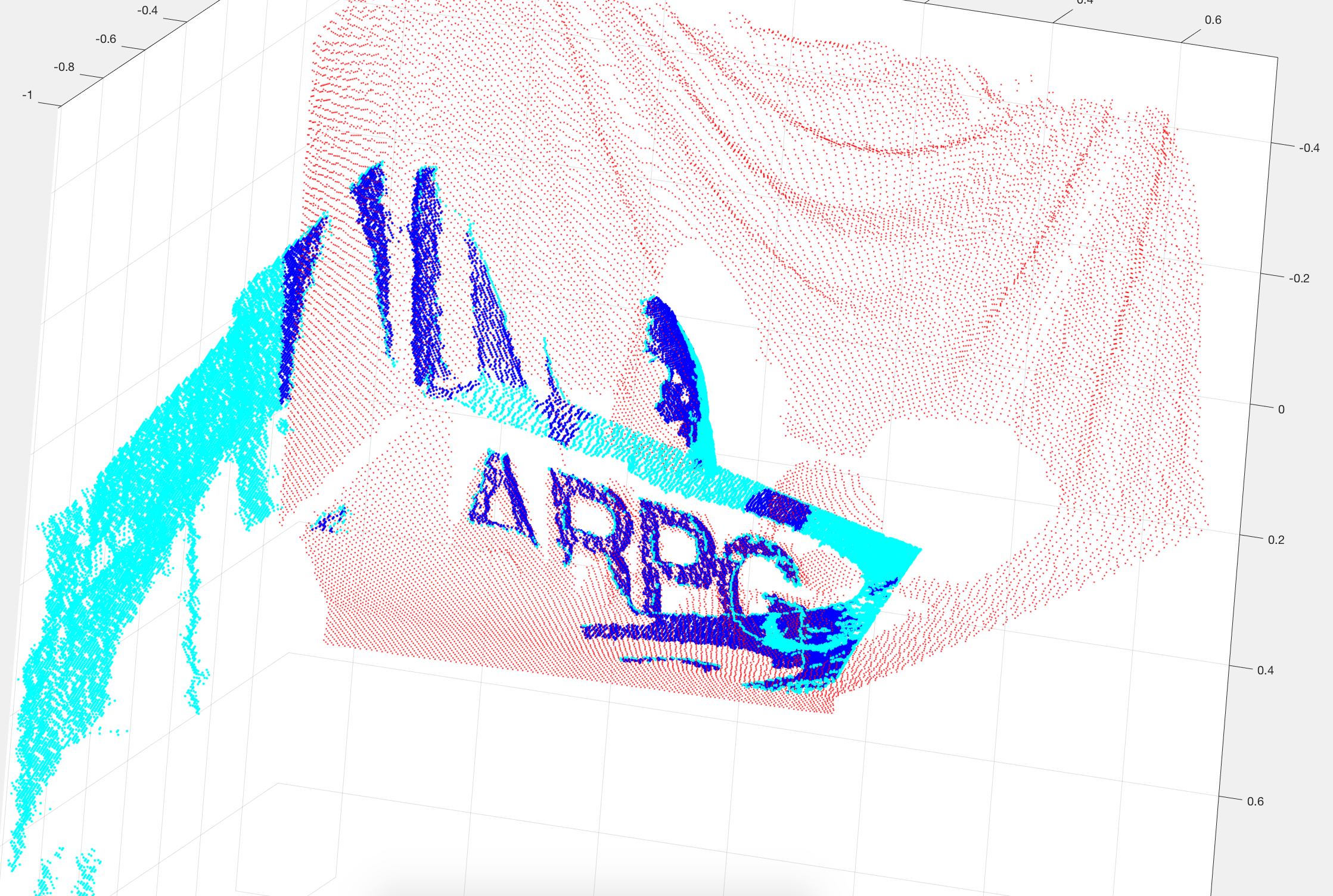}
    \caption{A successful registration with only 36\% overlap. The red points
    are the fixed cloud (which have been downsampled in this view), to which the
    blue points have been aligned. The dark blue points are inliers and the
    light blue are outliers.}\label{clouds}
  \end{center}
\end{figure}

ICP attempts to determine the optimal transformation to align two point clouds.
The algorithm recovers the transformation that moves the ``free'' point cloud
onto the ``fixed'' point cloud. ICP proceeds by iteratively:
\begin{enumerate}
  \item finding the closest point in the fixed cloud to each point in the free
    cloud;
  \item discarding some of these matches as outliers; and
  \item computing and applying the Euclidean transform that optimally aligns
    the remaining points, minimizing some measure of the distances to the
    nearest point found in step 1
\end{enumerate}
until convergence. The present work focuses on step 2 of the above sequence.

Variations on step 3, depending on specifically what error metric is
minimized~\cite{icpvariants}, are explored further in
Section~\ref{related}. There are closed-form
solutions~\cite{eggert1997estimating} that minimize the point-to-point distance,
including using the singular value decomposition~\cite{svdregistration} and the
dual number quaternion method~\cite{dualquat}. The present work minimizes the
sum of squared point-to-point distances, for simplicity, although it is
fundamentally independent of choice of error metric.

A crucial step in ICP is the rejection of outliers, generally resulting from
non-overlapping volumes of space between two measurements. The original ICP
formulation~\cite{icp_besl} did not discard any points and simply incurred
error for each outlier. A proliferation of
strategies have been proposed for discarding outliers
\cite{turkzippered,masudareg,trimmedicp,fusiello2002model,icp_zhang,enqvist2009optimal}.
In the present work we propose an alternative method where the distribution of
distances is modeled as a mixture of two Gaussians, one for inliers and one for
outliers.  Using a technique from image segmentation, a point's inlier or
outlier state is influenced by the state of its neighbors through a hidden
Markov random field. The EM algorithm, with the use of a mean field
approximation, allows for inference of the hidden state.

\section{Related Work}\label{related}

Several attempts have been made to improve registration performance in the
low-overlap regime. The hybrid genetic/hill-climbing algorithm
of~\cite{silva2003low,silva2005precision} shows success with overlaps down to
55\%. Good low-overlap performance is claimed in~\cite{struct_reg}, which
defines a ``direction angle'' on points and then aligns clouds in rotation using
a histogram of these, and in translation using correlations of 2-D projections.
Rotational alignment is recovered using extended Gaussian images
in~\cite{makadia2006fully}, and refined with ICP, showing success with overlap
as low as 45\%.

Use of robust statistics or error metrics besides sum of squared point-to-point
distance has made ICP more robust to outliers. The
point-to-plane~\cite{pointtoplane} metric, where the distance is measured as it
is projected onto the surfance normal of the fixed point, is frequently used and
has been shown to be more robust in the case of limited
overlap~\cite{salvi2007review}. Color information can also be
exploited~\cite{coloricp,color_icp}. Sparse norms are used within ICP
in~\cite{bouaziz2013sparse}, which can be sped up using simulated
annealing~\cite{mavridis2015efficient}.

Improvements to ICP have been achieved through better rejection of outlier
correspondences.~\cite{turkzippered} applies a user-defined distance threshold,
where matches beyond this threshold are discarded. Trimmed ICP uses fixed
fraction of point correspondences (typically 90\%) with smallest
residuals~\cite{trimmedicp}. Another approach uses a distance threshold of the
mean distance plus 2.5 times the standard deviation of the
distances~\cite{masudareg}.  The X84 criterion is similar, but uses the more
robust median absolute deviation (MAD) in place of the standard deviation, and
so rejects residuals more than 5.2 MADs above the
median~\cite{fusiello2002model}.~\cite{icp_zhang} presents an adaptive threshold
which is tuned with a distance parameter which does not have direct physical
significance. Yet another method~\cite{enqvist2009optimal} leverages the fact
that distances between corresponding points within a cloud will be invariant
under rigid motion and finds the largest set of consistent correspondences to
identify inliers. Fractional ICP dynamically adapts the fraction of point
correspondences to be used~\cite{phillips2007outlier}, although this method
explicitly assumes a large fraction of overlap, with a penalty for smaller
overlap fractions. In~\cite{masuda1995robust}, points are classified as inliers
or one of three classes of outliers: occluded, unpaired (outside the frame), or
outliers (sensor noise). EM-ICP~\cite{granger2002multi,hermans2011robust} treats
correspondence as a hidden variable and then computes assignment probabilities
in the E-step, and the implied rigid transformation in the M-step, however, it
is still assumed that every point in the free cloud corresponds to some point in
the fixed cloud.

Since ICP requires good initialization, a significant body of work has been
devoted to achieving global registration or finding an approximate alignment
from any initial state, which is then refined with ICP.\@
In~\cite{aiger20084,mellado2014super}, an initial alignment is found by matching
sets of 4 coplanar points, using ratios invariant to match sets; good
performance with low overlap is claimed. Fast global registration between two or
more point clouds is achieved in~\cite{zhou2016fast} by finding correspondences
once based on point features, and then using robust estimators to ignore
spurious correspondences. Kernel correlation~\cite{tsin2004correlation}
registers point clouds by minimizing the entropy of the resulting joint
cloud.~\cite{yousrya3d} minimizes the Hausdorff distance of several descriptors
using particle swarm optimization to find the globally optimal alignment.
LM-ICP~\cite{robust_2d_3d} uses Levenberg-Marquardt to directly minimize the
registration error over transformations and closest-point distances; derivatives
for closest-point distances are precomputed with finite differences of a
discretized distance transform.

Branch-and-bound algorithms can guarantee global optimality, and several
variations have been applied to the registration
problem~\cite{olsson2009branch,papazov2011stochastic}.
A branch-and-bound search for an initial alignment can be executed over
rotations assuming the translation is known~\cite{parra2014fast,li20073d}, or
over both rotations and translations~\cite{yang2016go}. Similarly,
branch-and-bound has been employed for point or feature correspondences, such
as~\cite{gelfand2005robust}, based on curvature features. Calculating bounds can
be computationally intensive, and so finding bounds that are easily calculated
offers appreciable speed-ups. For instance,~\cite{straubefficient} uses
distributions of surface normals for rotation and Gaussian mixtures for
translation.

Good feature descriptors for points in point clouds can help achieve global
registration efficiently by finding corresponding points independent of their
initial positions. Several feature descriptors are reviewed
in~\cite{guo2016comprehensive}. Deep neural network auto-encoders also have been
used to provide descriptors~\cite{dnn_reg}.

Finally, a variety of algorithms seek to register point clouds or shapes by
representing them as probability densities. Several methods for generating and
aligning these densities have been presented.
NDT~\cite{biber2003normal,magnusson2007scan,magnusson2009evaluation} uses a grid
of normal distributions to describe the probability of observing a point at each
possible location. Gaussian mixture models are frequently used to represent
point clouds~\cite{jian2005robust,campbell2015adaptive}.
Similarly, in~\cite{bylow2013real} a depth image is registered to a truncated
signed distance function representation of the target geometry. Non-rigid
transforms are estimated in~\cite{myronenko2010point}, using a GMM model for one
cloud and selecting the transformation that maximizes the likelihood of
observing the other cloud in that model.

Few of the above methods specifically address the problem of aligning clouds
with little overlap---more often, it is assumed that only a small portion of the
cloud will comprise unmatched points. By using an appropriate probabilistic
model, this assumption need not be made, and the resulting method works equally
well with high and low overlap.

\section{Methodology}
Let $i$ be a pixel position in the depth map generating the free cloud, and $B_i
\in \R^4$ be a point in the free cloud in homogenous coordinates. Let the fixed
points be $C_j \in \R^4$, again in homogenous coordinates. Note that the number
of fixed points need not equal the number of free points.

In the first step of the ICP iteration we find,
  \[ I_i = \argmin_{j} \lVert B_i - C_j \rVert. \]
We also save the associated distance,
  \[ Y_i = \min_{j} \lVert B_i - C_j \rVert = \lVert B_i - C_{I_i} \rVert. \]
The collected $Y_i$ random variables will be denoted $\bm{Y}$, with specific
realizations denoted $y_i$ and $\bm{y}$, respectively.

The second step in the iteration makes use of these distance values to determine
which point correspondences to consider inliers for the localization step,
  \[ S = \{i | \phi(i, Y_i)\}, \]
with some decision algorithm $\phi$.

Finally, the third step calculates the transformation,
  \[ T^* = \argmin_{T} \sum_{i \in S} f(TB_i, C_{I_i}) \]
with $T \in \mathbb{SE}(3)$ and $f$ an error metric. For the point-to-point
least-squares metric, which is used in this work, $f(TB_i, C_{I_i}) = {\lVert
TB_i - C_{I_i} \rVert}_2^2$.

For many sources of depth data, such as depth cameras or stereo matching, the
points have an underlying \mbox{2-D} lattice structure---the pixel grid. The
pixel grid can be used to define neighbor relationships, so that a pixel not
on the border of an image has four neighbors: up, down, left, and right. This
method exploits these neighbor relations to model the distribution of the
closest point distances $\bm{Y}$.

Given observed distances $\bm{y}$, we wish to decide which points are inliers.
Our prior beliefs are:
\begin{itemize}
  \item inliers will, on average, lie closer to their respective closest point
    than outliers; and
  \item pixel neighbors of inliers are likely inliers, and pixel neighbors of
    outliers are likely outliers.
\end{itemize}
To capture these priors, we model the distribution of $\bm{Y}$ as a mixture
of two Gaussians, one for inliers and one for outliers, where a point's mixture
membership is dependent on its four nearest pixel neighbors. That is, we
capture the second prior using a hidden Markov random field on the
inlier/outlier state of a point. The graphical model is shown in Fig.~\ref{pgm}.
The distribution of $\bm{Y}$ is conditionally dependent on the hidden
distribution $\bm{Z}$ which is Gibbs distributed. The maximum likelihood
model parameters and hidden state can be estimated using the EM algorithm, with
a mean field approximation for $\bm{Z}$, which does not require any Monte
Carlo methods. This method is very similar to the image segmentation model
presented in~\cite{kato2006markov,besag1986statistical,em_hmrf_celeux}, which
we follow closely, specializing its derivation to this application.

\subsection{Hidden Markov random field model}
Let $Z_i \in \{\pm1\}$ represent the inlier/outlier state of the point generated
from pixel $i$, with $-1$ indicating an outlier, and $+1$ indicating an inlier.
Let $\bm Z$ represent the collection of all $Z_i$ states, and $\bm z$ represent
a realization of $\bm Z$. Let $i \sim i'$ indicate that two pixels $i$ and $i'$
are neighbors. $\bm Z$ is a Markov random field, and is Gibbs distributed.

A Bayesian network can be factored into conditional distributions, which can be
calculated directly. However, a Markov random field has undirected edges and so
the distribution is calculated based on an energy function, $H(\bm{Z})$, from
which the relative probability of different configurations can be calculated.

The Gibbs distribution is calculated based on the energy of a given
configuration $\bm z$, appropriately normalized,
  \[ P_G(\bm{Z}) = W^{-1} \exp(-H(\bm{Z})), \]
where $W$ is a normalization term called the ``partition function'',
  \[ W = \sum_\mathbf{z} \exp(-H(\bm{z})). \]

We use the energy function, $H(\bm{z}) = -\beta \sum_{i \sim i'} z_i z_{i'}$,
where $\beta \geq 0$ is a parameter controlling the interaction strength ($\beta
< 0$ represents systems where we expect neighbors to be \emph{dis}similar, and
when $\beta = 0$ there is no interaction). Note that calculation of $W$, and
therefore exact calculation of $P_G(\bm{z})$, requires a sum over all possible
configurations $\bm z$, so is exponential in the number of pixels. For any
nontrivial point cloud this is intractable. We avoid this issue with the mean
field approximation, described below.

We assume $Y_i | Z_i \sim N(y_i; \mu_{z_i}, \sigma_{z_i})$, so the complete
distribution, then, is
\[f(\bm{Y}, \bm{Z} | \beta, \theta) = P_G(\bm{Z} | \beta) \prod_i N(y_i |
    \mu_{z_i}, \sigma_{z_i})\]
with parameters $\{\beta, \mu_{-1}, \sigma_{-1}, \mu_{+1}, \sigma_{+1}\}$. We
will use $\theta$ to represent all Gaussian component parameters, that is,
$\theta = \{\mu_{-1}, \sigma_{-1}, \mu_{+1}, \sigma_{+1}\}$.

The mean field approximation assumes a fixed configuration $\bm{\tilde{z}}
= \E[P_G(\bm{Z}|\beta)]$, and approximates the Gibbs distribution with
independent components conditioned on this fixed configuration,
\[ P_G(\bm{Z}|\beta) \approx \prod_i P(Z_i | \beta, \bm{\tilde{z}}). \]
As the components are independent, it is no longer necessary to exhaust over
all possible $\bm{z}$ configurations. Note that while $z_i \in \{-1, +1\}$,
the mean field can take intermediate values, $\tilde{z}_i \in [-1, +1]$. The
components also depend only on local information,
\[ P(z_i | \beta, \bm{\tilde{z}}) = \frac{\exp\left(\beta \sum_{i \sim i'} z_i
  \tilde{z}_{i'}\right)}{ \exp\left(\beta \sum_{i \sim i'} (+1)
\tilde{z}_{i'}\right) + \exp\left(\beta \sum_{i \sim i'} (-1)
\tilde{z}_{i'}\right)}. \]

The final complete likelihood with the mean field approximation is then,
\[f(\bm{Y}, \bm{Z} | \beta, \theta) = P(z_i | \beta, \bm{\tilde z}) N(y_i | \mu_{z_i}, \sigma_{z_i}),\]
and the log likelihood is,
\begin{multline*}
  \log f(\bm{Y}, \bm{Z} | \beta, \theta) = \\
  \sum_i \left( \beta \sum_{i \sim i'} z_i \tilde{z}_{i'} -  \right.
  \log \left(\exp\left(\beta \sum_{i \sim i'} (+1) \tilde{z}_{i'}\right) +
    \exp\left(\beta \sum_{i \sim i'} (-1) \tilde{z}_{i'}\right) \right) \\
  \left. - \frac{1}{2} \sigma_{z_i} - \frac{{(y_i - \mu_{z_i})}^2}{2
  \sigma_{z_i}^2}\right) + \text{const}.
\end{multline*}

\subsection{Applying the EM algorithm}
With this approximation, the EM algorithm~\cite{em} can be adapted to find the
maximum likelihood estimate of the parameters $\theta$ and the hidden field. The
$\beta$ value could be estimated as well, but we simply use the value 2, which
gives satisfying results. Initially, we assume that the 10\% of points with the
greatest distance to their corresponding closest point are outliers and so set
$\bm{\tilde{z}}^{(0)}$ accordingly. Similarly, $\theta^{(0)}$ is
initialized with the means and standard deviations of the two sets. The EM
iteration then proceeds as follows,
\begin{itemize}
  \item \emph{E-step:}
    \[\bm{\tilde{z}}^{(q+1)} \coloneqq \E_\text{mf}[\bm{Z} | \bm{y},
        \bm{\tilde{z}}^{(q)}, \theta^{(q)}];\]
  \item \emph{M-step:}
    \[\theta^{(q+1)} \coloneqq \argmax_\theta f(\bm{Y}, \bm{Z}|\beta, \theta) |
       \bm{y}, \bm{\tilde{z}}^{(q+1)}.\]
\end{itemize}
We integrate this into the ICP algorithm by using the resulting
$\bm{\tilde{z}}^{(q+1)}$ values to select inliers. We consider EM to have
converged when all corresponding values in $\bm{\tilde{z}}^{(q+1)}$ and
$\bm{\tilde{z}}^{(q)}$ have the same sign. Due to the potential for oscillatory
behavior, we also compare $\bm{\tilde{z}}^{(q+1)}$ and $\bm{\tilde{z}}^{(q-1)}$.

Ideally, we would run EM to convergence within each ICP step. However, we limit
the number of EM iterations to ensure adequate performance. Since the initial
assumptions may be poor, we allow more EM iterations before calculating and
applying the first transformation. Within each ICP step, EM is limited to 20
iterations, but initially it is allowed to run up to 600 steps, which allowed
for convergence in most cases. The full procedure is shown in
Algorithm~\ref{fullalg}.

\subsubsection{The E-step}
With the mean field approximation, this step is similar to the E-step of a
simple Gaussian mixture model, although each point has its own prior
distribution on component membership. The update is independent for each point,
and is calculated
\[ \tilde{z}_i^{(q+1)} = \frac{\sum_{z_i} z_i
      \exp\left(\beta\sum_{i \sim i'} z_i \tilde{z}_{i'}^{(q)} - \log
\sigma_{z_i}- 0.5 \sigma_{z_i}^{-2} {(y_i - % chktex 9
\mu_{z_i})}^2\right)}{\sum_{z_i} \exp\left(\beta % chktex 9
\sum_{i \sim i'} z_i \tilde{z}_{i'}^{(q)} - \log \sigma_{z_i}
- 0.5 \sigma_{z_i}^{-2} {(y_i - \mu_{z_i})}^2\right) %chktex 9
}. \]

Note that~\cite{em_hmrf_celeux} recommends a sequential update, but this is
simultaneous, which is both embarrassingly parallel and easier to implement.

\subsubsection{The M-step}
The $\tilde{z}_i^{(q+1)}$ values calculated in the E-step are simply linear
transformations of the current estimate of $P(Z_i = z_i)$,
\[ P^{(q+1)}(Z_i = z_i) = \frac{1 + z_i \tilde{z}_i^{(q+1)}}{2}, \]
so the M-step update is simply,
\begin{align*}
  \mu_{z_i}^{(q+1)} &= \frac{\sum_i P^{(q+1)}(Z_i = z_i) Y_i}{\sum_i
      P^{(q+1)}(Z_i = z_i)} \\
  \sigma_{z_i}^{(q+1)} &= \frac{\sum_i P^{(q+1)}(Z_i = z_i) {(Y_i -
      \mu_{z_i}^{(q+1)})}^2}{\sum_i P^{(q+1)}(Z_i = z_i)}.
\end{align*}

\begin{algorithm}[t]
  \KwIn{$T_\text{init}$, $\bm{B}$, $\bm{C}$, $\beta$, thresholds}

  KDtree = BuildKdTree$(\bm{C})$

  $T = T_\text{init}$

  $\bm{B} = T \times \bm{B}$

  $\bm{I}, \bm{Y}$ = KDtree.NearestNeighbors$(\bm{B})$

  Initialize $\bm{z}$ to 1 except for highest 10\% of $\bm{Y}$, which
  are initialized to -1.

  \Do{some $\bm{z}$ value changed sign and iters $<$ 600}{
    $\theta$ = M-step$(\bm{Y}, \bm{z})$

    $\bm{z}$ = E-step$(\bm{Y}, \theta, \beta)$
  }

  \Do{convergence thresholds not met and iters $<$ 50}{

    \Do{some $\bm z$ value changed sign and iters $<$ 20}{
      $\theta$ = M-step$(\bm{Y}, \bm{z})$

      $\bm{z}$ = E-step$(\bm{Y}, \theta, \beta)$
    }

    $T_\text{step}$ = localize$(\bm{B}, \bm{C}, \bm{z})$

    $T = T_\text{step} \times T$

    $\bm{B} = T_\text{step} \times \bm{B}$

    $\bm{I}, \bm{Y}$ = KDtree.NearestNeighbors$(\bm{B})$
  }
  \Return{T}
  \caption{The full HMRF ICP algorithm}\label{fullalg}
\end{algorithm}

\section{Experiments}
\begin{figure}[t]
  \begin{center}
    \includegraphics[width=0.45\textwidth]{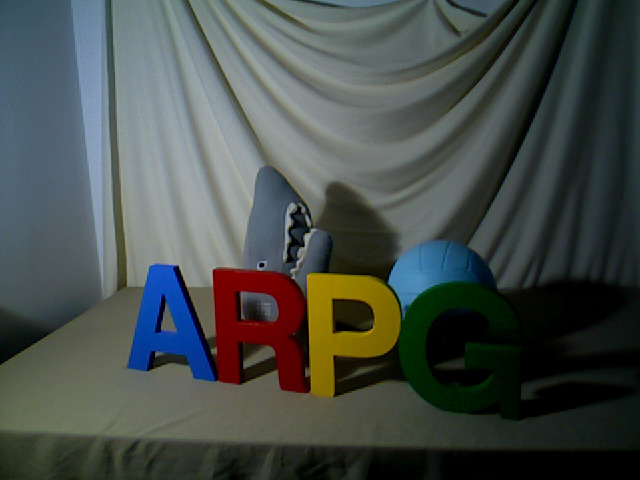}
    \includegraphics[width=0.45\textwidth]{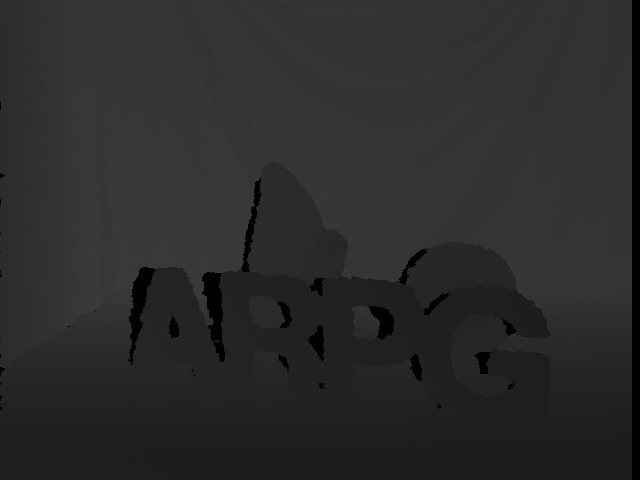}
    \includegraphics[width=0.45\textwidth]{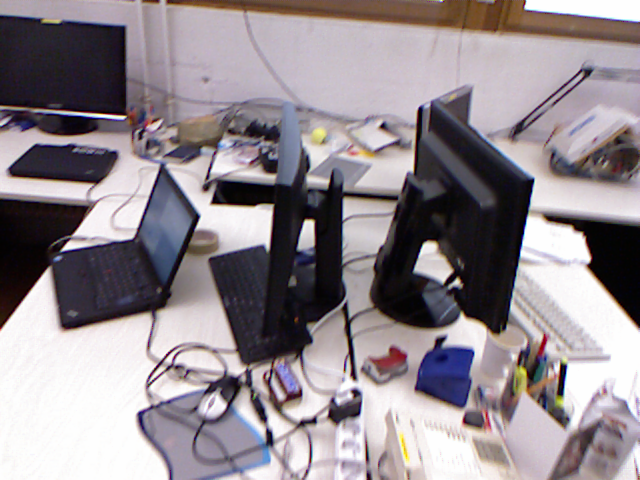}
    \includegraphics[width=0.45\textwidth]{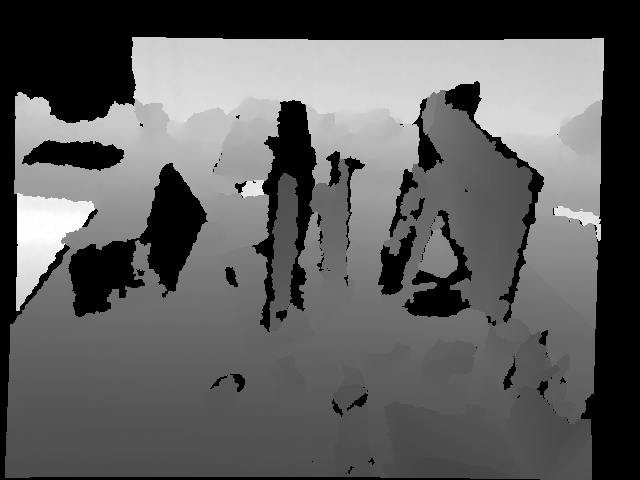}
    \caption{Fixed frames, RGB (not used in alignment) and depth
    maps.}\label{ref_frames}
  \end{center}
\end{figure}

The method is compared with five other outlier rejection methods: no outlier
rejection, keeping 90\% of the points with smallest residuals, keeping points
whose residuals were less than 2.5 standard deviations above the mean, keeping
points whose residuals were less than 5.2 median absolute deviations above the
median (the so-called ``X84'' criterion), and the dynamic threshold
from~\cite{icp_zhang}. These are referred to as ``all'', ``percent'', ``sigma'',
``X84'', and ``dynamic'', respectively.

The method is applied to two datasets: the shark sequence is a tabletop scene,
taken by an Asus Xtion Pro; the desk sequence is a publicly available RGB-D SLAM
dataset~\cite{sturm12iros}. Ground truth poses are known for both sequences. The
sensor in the desk sequence is tracked using an external motion tracking system,
operating at 100Hz, from which poses at shutter timestamps are interpolated. The
poses in the shark sequence are those estimated in the tracking stage of
InfiniTAM\cite{kahler2015very} during the reconstruction of the scene. The
intrinsic camera calibration for the desk sequence is provided with the data and
the Xtion Pro was calibrated so that the depth maps could be unprojected
appropriately and converted to point clouds.

In each case, one frame is chosen as the reference frame, shown in
Fig.~\ref{ref_frames}, and then a number of other frames with varying fractions
of overlap are aligned to it. The remaining frames are a random
sample stratified to include a variety of overlap ratios the with fixed frame.
The overlap with the fixed frame is estimated using the ground truth poses and
the distance to each point's nearest neighbor in its own cloud versus in the
fixed cloud.

In the full shark sequence, the smallest overlap is about 35\%, and the largest
nearly 100\%. The sample of 40 frames is selected with 5 frames from each
decile (30\% to 40\%, 40\% to 50\%, etc.), and 5 frames selected at random from
all the frames.

The desk sequence includes frames that do not overlap at all with the reference
frame; these frames are discarded. However, the sample includes frames
with less than 1\% overlap. The sample is again a stratified random sample,
with 5 frames selected from each decile, for a total of 50 frame.

\section{Results}
For each free frame, 16 initializations were tested. These were generated by
aligning each frame to the fixed frame using the ground truth poses, and then
perturbing this alignment by a rotation of $\pi/30$ around each of 16 random
axes, centered at the cloud centroid. The same 16 axes were used for all frames.
All experiments were executed in MATLAB on a workstation with 8
Intel\textsuperscript{\textregistered} Xeon\textsuperscript{\textregistered}
E5620 CPUs at 2.40GHz, and with 48 Gb RAM.\@ All code can be found at
\url{https://github.com/JStech/ICP}.

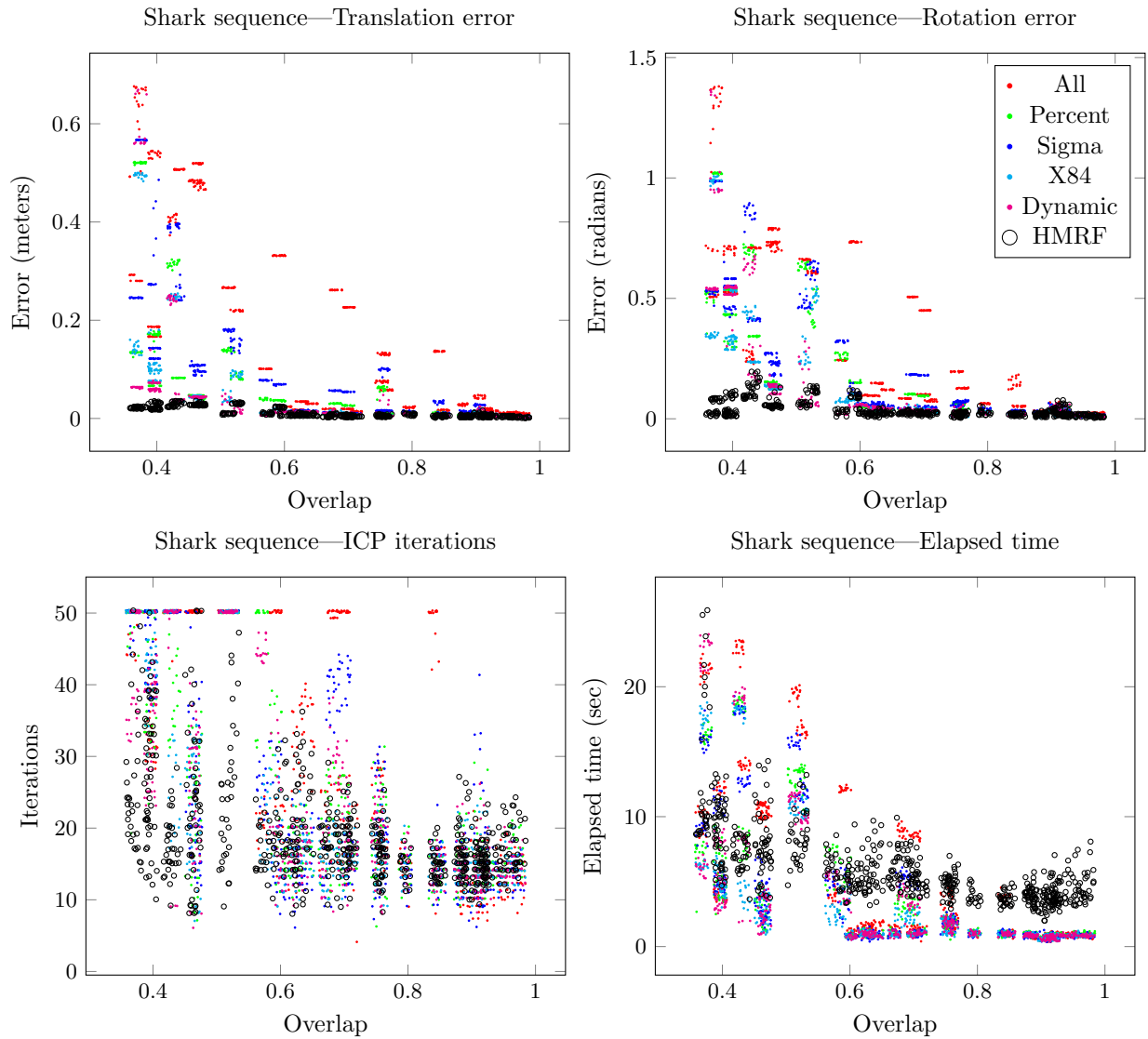
\begin{figure*}
  \begin{center}
    \makebox[\linewidth][c]{
      \begin{tikzpicture}
        \begin{axis}[title=Shark sequence---Translation error,
          xlabel = Overlap,
          ylabel = Error (meters),
          ]
          \addplot[red, mark size=0.3pt] table [x=overlap, y=all_t_err, col sep=comma] {results_shark.csv};
          \addplot[green, mark size=0.3pt] table [x=overlap, y=pct_t_err, col sep=comma] {results_shark.csv};
          \addplot[blue, mark size=0.3pt] table [x=overlap, y=sigma_t_err, col sep=comma] {results_shark.csv};
          \addplot[cyan, mark size=0.3pt] table [x=overlap, y=x84_t_err, col sep=comma] {results_shark.csv};
          \addplot[magenta, mark size=0.3pt] table [x=overlap, y=dynamic_t_err, col sep=comma] {results_shark.csv};
          \addplot[black, mark = o, mark size=1pt] table [x=overlap, y=hmrf_t_err, col sep=comma] {results_shark.csv};
        \end{axis}
      \end{tikzpicture}
    \begin{tikzpicture}
      \begin{axis}[title=Shark sequence---Rotation error,
        xlabel = Overlap,
        ylabel = Error (radians),
        legend pos = north east,
        legend image post style={scale=3},
        ]
        \addplot[red, mark size=0.3pt] table [x=overlap, y=all_r_err, col sep=comma] {results_shark.csv};
        \addplot[green, mark size=0.3pt] table [x=overlap, y=pct_r_err, col sep=comma] {results_shark.csv};
        \addplot[blue, mark size=0.3pt] table [x=overlap, y=sigma_r_err, col sep=comma] {results_shark.csv};
        \addplot[cyan, mark size=0.3pt] table [x=overlap, y=x84_r_err, col sep=comma] {results_shark.csv};
        \addplot[magenta, mark size=0.3pt] table [x=overlap, y=dynamic_r_err, col sep=comma] {results_shark.csv};
        \addplot[black, mark = o, mark size=1pt] table [x=overlap, y=hmrf_r_err, col sep=comma] {results_shark.csv};
        \legend{All, Percent, Sigma, X84, Dynamic, HMRF}
      \end{axis}
    \end{tikzpicture}
  }

    \makebox[\linewidth][c]{
      \begin{tikzpicture}
        \begin{axis}[title=Shark sequence---ICP iterations,
          xlabel = Overlap,
          ylabel = Iterations,
          jittery=0.4
          ]
          \addplot[red, mark size=0.3pt] table [x=overlap, y=all_iters, col sep=comma] {results_shark.csv};
          \addplot[green, mark size=0.3pt] table [x=overlap, y=pct_iters, col sep=comma] {results_shark.csv};
          \addplot[blue, mark size=0.3pt] table [x=overlap, y=sigma_iters, col sep=comma] {results_shark.csv};
          \addplot[cyan, mark size=0.3pt] table [x=overlap, y=x84_iters, col sep=comma] {results_shark.csv};
          \addplot[magenta, mark size=0.3pt] table [x=overlap, y=dynamic_iters, col sep=comma] {results_shark.csv};
          \addplot[black, mark = o, mark size=1pt] table [x=overlap, y=hmrf_iters, col sep=comma] {results_shark.csv};
        \end{axis}
      \end{tikzpicture}
    \begin{tikzpicture}
      \begin{axis}[title=Shark sequence---Elapsed time,
        xlabel = Overlap,
        ylabel = Elapsed time (sec),
        ]
        \addplot[red, mark size=0.3pt] table [x=overlap, y=all_time, col sep=comma] {results_shark.csv};
        \addplot[green, mark size=0.3pt] table [x=overlap, y=pct_time, col sep=comma] {results_shark.csv};
        \addplot[blue, mark size=0.3pt] table [x=overlap, y=sigma_time, col sep=comma] {results_shark.csv};
        \addplot[cyan, mark size=0.3pt] table [x=overlap, y=x84_time, col sep=comma] {results_shark.csv};
        \addplot[magenta, mark size=0.3pt] table [x=overlap, y=dynamic_time, col sep=comma] {results_shark.csv};
        \addplot[black, mark = o, mark size=1pt] table [x=overlap, y=hmrf_time, col sep=comma] {results_shark.csv};
      \end{axis}
    \end{tikzpicture}
  }
    \caption{Shark sequence results}\label{shark_res}
  \end{center}
\end{figure*}

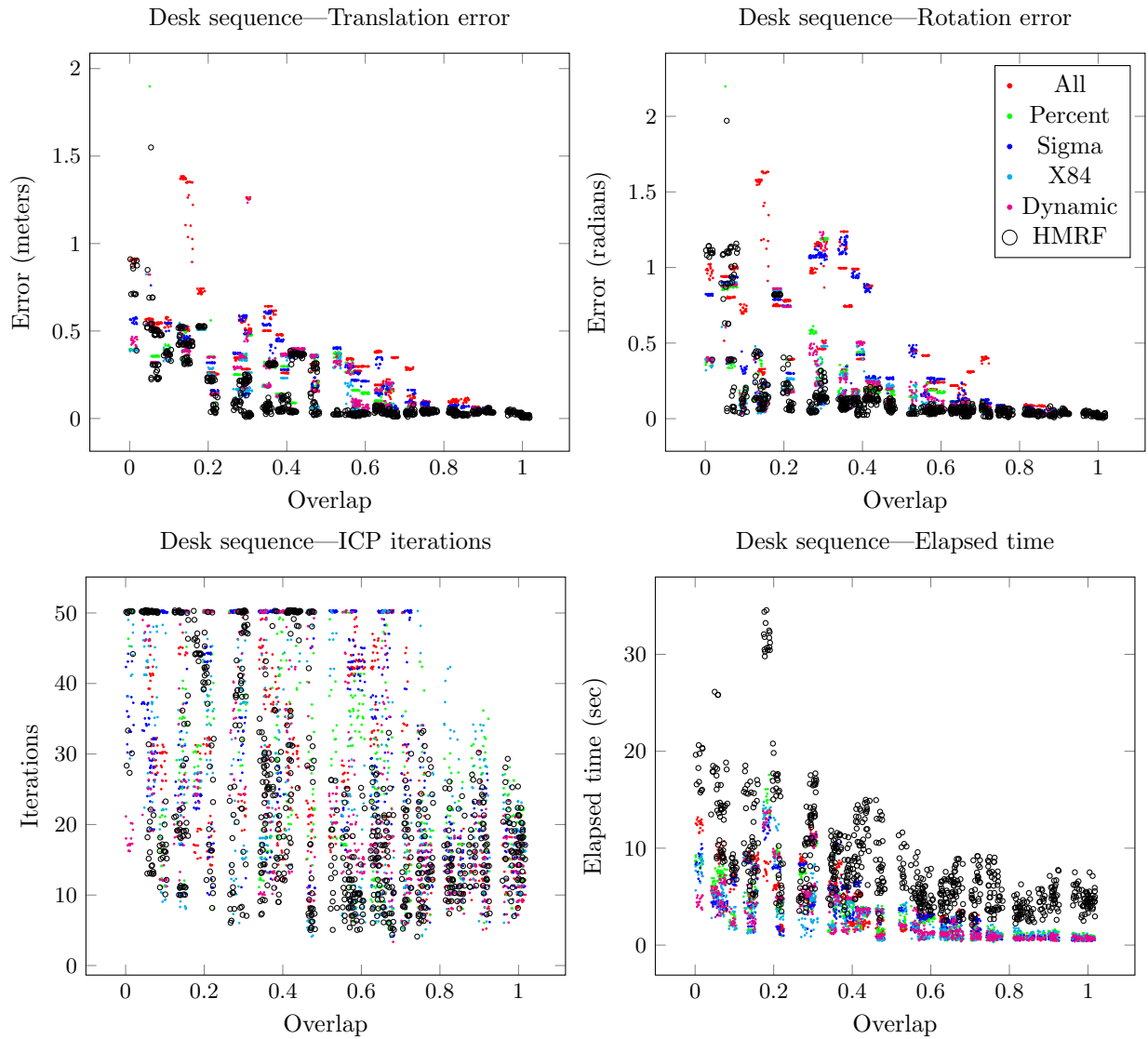
\begin{figure*}
  \begin{center}
    \makebox[\linewidth][c]{
      \begin{tikzpicture}
        \begin{axis}[title=Desk sequence---Translation error,
          xlabel = Overlap,
          ylabel = Error (meters),
          ]
          \addplot[red, mark size=0.3pt] table [x=overlap, y=all_t_err, col sep=comma] {results_desk.csv};
          \addplot[green, mark size=0.3pt] table [x=overlap, y=pct_t_err, col sep=comma] {results_desk.csv};
          \addplot[blue, mark size=0.3pt] table [x=overlap, y=sigma_t_err, col sep=comma] {results_desk.csv};
          \addplot[cyan, mark size=0.3pt] table [x=overlap, y=x84_t_err, col sep=comma] {results_desk.csv};
          \addplot[magenta, mark size=0.3pt] table [x=overlap, y=dynamic_t_err, col sep=comma] {results_desk.csv};
          \addplot[black, mark = o, mark size=1pt] table [x=overlap, y=hmrf_t_err, col sep=comma] {results_desk.csv};
        \end{axis}
      \end{tikzpicture}
      \begin{tikzpicture}
        \begin{axis}[title=Desk sequence---Rotation error,
          xlabel = Overlap,
          ylabel = Error (radians),
          legend pos = north east,
          legend image post style={scale=3},
          ]
          \addplot[red, mark size=0.3pt] table [x=overlap, y=all_r_err, col sep=comma] {results_desk.csv};
          \addplot[green, mark size=0.3pt] table [x=overlap, y=pct_r_err, col sep=comma] {results_desk.csv};
          \addplot[blue, mark size=0.3pt] table [x=overlap, y=sigma_r_err, col sep=comma] {results_desk.csv};
          \addplot[cyan, mark size=0.3pt] table [x=overlap, y=x84_r_err, col sep=comma] {results_desk.csv};
          \addplot[magenta, mark size=0.3pt] table [x=overlap, y=dynamic_r_err, col sep=comma] {results_desk.csv};
          \addplot[black, mark = o, mark size=1pt] table [x=overlap, y=hmrf_r_err, col sep=comma] {results_desk.csv};
          \legend{All, Percent, Sigma, X84, Dynamic, HMRF}
        \end{axis}
      \end{tikzpicture}
    }

    \makebox[\linewidth][c]{
      \begin{tikzpicture}
        \begin{axis}[title=Desk sequence---ICP iterations,
          xlabel = Overlap,
          ylabel = Iterations,
          jittery=0.4
          ]
          \addplot[red, mark size=0.3pt] table [x=overlap, y=all_iters, col sep=comma] {results_desk.csv};
          \addplot[green, mark size=0.3pt] table [x=overlap, y=pct_iters, col sep=comma] {results_desk.csv};
          \addplot[blue, mark size=0.3pt] table [x=overlap, y=sigma_iters, col sep=comma] {results_desk.csv};
          \addplot[cyan, mark size=0.3pt] table [x=overlap, y=x84_iters, col sep=comma] {results_desk.csv};
          \addplot[magenta, mark size=0.3pt] table [x=overlap, y=dynamic_iters, col sep=comma] {results_desk.csv};
          \addplot[black, mark = o, mark size=1pt] table [x=overlap, y=hmrf_iters, col sep=comma] {results_desk.csv};
        \end{axis}
      \end{tikzpicture}
      \begin{tikzpicture}
        \begin{axis}[title=Desk sequence---Elapsed time,
          xlabel = Overlap,
          ylabel = Elapsed time (sec),
          ]
          \addplot[red, mark size=0.3pt] table [x=overlap, y=all_time, col sep=comma] {results_desk.csv};
          \addplot[green, mark size=0.3pt] table [x=overlap, y=pct_time, col sep=comma] {results_desk.csv};
          \addplot[blue, mark size=0.3pt] table [x=overlap, y=sigma_time, col sep=comma] {results_desk.csv};
          \addplot[cyan, mark size=0.3pt] table [x=overlap, y=x84_time, col sep=comma] {results_desk.csv};
          \addplot[magenta, mark size=0.3pt] table [x=overlap, y=dynamic_time, col sep=comma] {results_desk.csv};
          \addplot[black, mark = o, mark size=1pt] table [x=overlap, y=hmrf_time, col sep=comma] {results_desk.csv};
        \end{axis}
      \end{tikzpicture}
    }
    \caption{Desk sequence results}\label{desk_res}
  \end{center}
\end{figure*}

Figs.~\ref{shark_res} and~\ref{desk_res} show the results, plotted against
overlap fraction. The translation error is the norm of the error vector
(measured in meters). The rotation error is the angle in radians. Finally, the
number of ICP iterations until convergence, as well as total elapsed time are
also shown. A small amount of jitter has been added to the x-axis in all plots,
to allow for better visual inspection of the data. The plot of iterations
also has y-axis jitter.

\begin{figure*}
  \begin{center}
    \includegraphics[width=0.3\textwidth]{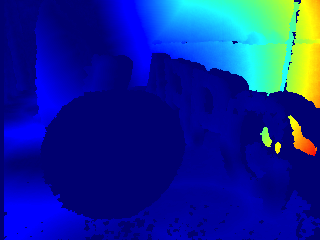}
    \includegraphics[width=0.3\textwidth]{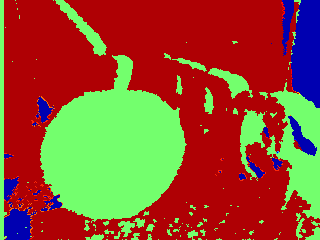}
    \includegraphics[width=0.3\textwidth]{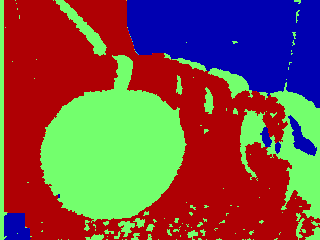}
    \caption{Example $\bm Y$, $\bm{\tilde{z}}^{(1)}$, and
    $\bm{\tilde{z}}^{(370)}$ (at convergence). This frame has 36\% overlap with
    the fixed frame. The left image shows the observed distance to the nearest
    fixed point in the initial configuration; red is farther, blue is nearer,
    and the darkest blue areas are unobserved. In the right two images, blue
    pixels are outliers and red pixels are inliers, green pixels are
    unobserved.}\label{ex_frame}
  \end{center}
\end{figure*}

At moderate to low overlaps, the HMRF method recovers a more accurate
transformation and does so in fewer iterations. To see why, consider the example
frame with 36\% overlap shown in Fig.~\ref{ex_frame}. The three images all
represent values before the first transformation is applied to the free cloud:
the first is the residual distance to the nearest fixed point, the second is the
initial setting of the $\bm{\tilde{z}}$ field, and the third is the converged
$\bm{\tilde{z}}$ field before the first iteration of ICP.\@ The HMRF model
flexibly adapts to the small proportion of inlier points, in particular it
eliminates outliers across the top of the image. The unobserved pixels occupy
31\% of the image. In the initial $\bm{\tilde{z}}$ field, 62\% of pixels are
considered inliers, and the remaining 7\% are considered outliers. After 370
initial EM iterations, the $\bm{\tilde{z}}$ field has converged, and now has
only 49\% inliers and 20\% outliers. By eliminating these outliers before the
first transformation is calculated, divergence from the nearby optimum is
avoided. The point clouds after final registration for this frame can be seen in
Fig.~\ref{clouds}.

HMRF ICP performs well for all frames in the shark sequence, with a maximum of
0.017 m translation error and 0.0776 radians rotation error for clouds with 60\%
or more overlap; the maximum error for all shark frames is 0.036 m and 0.196
radians. Furthermore, although HMRF ICP is slower than the other methods at high
overlap, it shows very little slowing as overlap decreases, so these low errors
were achieved faster than the poor results of the other methods.

The performance is not as good with the desk sequence. Three regimes are evident
in the results: above 50\% overlap, between 20\% and 50\%, and below 20\%.
The maximum error on frames with 50\% or more overlap is 0.077 m and 0.109
radians, and 0.386 m and 0.317 radians with 20\% or more overlap. Below 20\%
overlap, performance degrades severely, with maximum errors of 1.55 m and 1.97
radians.

An example alignment is shown at \url{https://youtu.be/w4eVOgd7Zes}.

\section{Conclusion}
The HMRF model for inliers and outliers in a depth map being aligned via ICP has
demonstrated advantages in the low overlap regime without sacrificing
performance at high overlap. The HMRF model describes observed inlier/outlier
behavior well, and so can adapt to the situation. This proves particularly
useful in the construction of models from 3-D scanner data, as fewer scans would
be required. There are numerous improvements that could be pursued for future
research. The Gaussian distribution was chosen for convenience, but it is
unlikely that residuals of either inliers or outliers are distributed normally.
It would be straightforward to use different distributions, which could be
selected by investigating the empirical inlier and outlier residual
distributions. This method also assumes an underlying grid topology for the free
cloud, which precludes it from use with a variety of sensors, such as rotating
LiDAR sensors. It could be extended to general, unstructured point clouds by
defining appropriate neighbor relations and energy function, $H(\bm z)$.
Finally, many of the improvements to ICP detailed in Section~\ref{related} could
be incorporated to expand the basin of convergence or otherwise improve
performance.

\section{Acknowledgements}
The authors thank Nisar Ahmed for his helpful insight. JS was supported by the
U.S.\ Department of Defense (DoD) through the National Defense Science \&
Engineering Graduate Fellowship (NDSEG) program while conducting this work. CH
was supported by DARPA award no.\ N65236--16--1--1000.

\bibliography{hmrf_icp}
\bibliographystyle{ieeetr}

\end{document}